\documentclass{article}
\usepackage{natbib}
\usepackage[utf8]{inputenc}
\usepackage[T1]{fontenc}    
\usepackage{hyperref}       
\usepackage{url}            
\usepackage{booktabs}       
\usepackage{amsfonts}       
\usepackage{nicefrac}       
\usepackage{microtype}      
\usepackage{lipsum}
\usepackage{comment}
\usepackage{subcaption}
\usepackage{amsmath}
\usepackage[linesnumbered,ruled,vlined]{algorithm2e}
\usepackage{graphicx}
\graphicspath{ {./images/} }

\title{Integrating Renewable Energy in Agriculture: A Deep Reinforcement Learning-based Approach}

\author{
 Abdul Wahid \\
  School of Computer Science\\
  University of Galway\\
  Galway, Ireland H91 FYH2 \\
  \texttt{abdul.wahid@universityofgalway.ie} \\
   \And
 Iias Faiud \\
  School of Computer Science\\
  University of Galway\\
  Galway, Ireland H91 FYH2 \\
  \texttt{i.faiud1@universityofgalway.ie} \\
  \And
 Karl Mason \\
  School of Computer Science\\
  University of Galway\\
  Galway, Ireland H91 FYH2 \\
  \texttt{karl.mason@universityofgalway.ie} \\
}
\begin{document}
\maketitle{
    \renewcommand{\thefootnote}{\fnsymbol{footnote}}
    \footnotetext{\textit{Proc. of the Deep Learning for Sustainable Precision Agriculture, ECML PKDD 2023, Marcello et al. (eds.), Sep 22, 2023, https://sites.google.com/view/dlspa-ecmlpkdd2023/. 2023.}}
}
\begin{abstract}
This article investigates the use of Deep Q-Networks (DQNs) to optimize decision-making for photovoltaic (PV) systems installations in the agriculture sector. The study develops a DQN framework to assist agricultural investors in making informed decisions considering factors such as installation budget, government incentives, energy requirements, system cost, and long-term benefits. By implementing a reward mechanism, the DQN learns to make data-driven decisions on PV integration. The analysis provides a comprehensive understanding of how DQNs can support investors in making decisions about PV installations in agriculture. This research has significant implications for promoting sustainable and efficient farming practices while also paving the way for future advancements in this field. By leveraging DQNs, agricultural investors can make optimized decisions that improve energy efficiency, reduce environmental impact, and enhance profitability. This study contributes to the advancement of PV integration in agriculture and encourages further innovation in this promising area.
\end{abstract}

\section{Introduction}
Photovoltaic (PV) system has gained significant traction in recent years as a sustainable and environmentally friendly alternative to traditional energy sources. Its potential to mitigate greenhouse gas emissions and combat climate change has made it a promising renewable energy solution. Recognizing the importance of renewable energy in achieving the Sustainable Development Goals (SDGs), particularly SDG 7, which aims to ensure universal access to affordable, reliable, sustainable, and modern energy services, the United Nations has emphasized the role of PV installations in promoting sustainable energy consumption and reducing reliance on non-renewable sources \cite{un2015sdg7}. By harnessing PV energy systems, this can contribute to the objective of expanding the use of sustainable energy and working towards a more sustainable and resilient future.

Conventional approaches to advising PV installations often rely on the expertise and experience of professionals or general rules of thumb that may not consider individual investor circumstances. Currently, there is a lack of dedicated tools specifically designed to address the intricacies of PV integration in agriculture. Therefore, there is a need for decision-making methods that can appropriately account for the complexities associated with this issue and provide investors with the necessary support to assess the viability and potential advantages of PV investments in agriculture.

One potential approach to address the challenges of decision-making in PV integration in agriculture is the utilization of Deep Q-Networks (DQNs). DQNs are a type of reinforcement learning algorithm that can learn to make decisions by maximizing a reward function. This approach has shown promising results in various domains, including robotics \cite{luo2023guiding}, gaming \cite{souchleris2023reinforcement}, and finance \cite{shavandi2022multi}. However, the extent of their applicability and potential in the agricultural sector is yet to be fully explored.  

This paper presents a deep reinforcement learning (DRL)-based decision-making system designed to provide recommendations on PV systems integration to agricultural investors. Our approach utilizes a deep neural network (DNN) to represent the decision policy, leveraging its powerful capabilities for learning and representing the optimal policy for suggesting PV systems installations. The main contributions of this paper are as follows: 
 
\begin{itemize}
    \item We propose an approach based on Deep Q-Networks (DQNs) for decision-making for PV systems integration in agriculture.
    \item This research provides valuable insights into the use of DQNs as a tool to assist investors in making informed decisions about installing PV systems.
    \item By leveraging the potential of DQNs, this article contributes to the advancement of the UN's sustainable energy goal by promoting informed decision-making in PV installations.
\end{itemize}
The paper comprises the following sections: Section 2 covers related work. Section 3 provides a background on reinforcement learning. Section 4 presents the DQN-based decision-making system for PV integration into agriculture. Section 5 discusses the experimental setup and results, and Section 6 concludes with a summary and future work.
\section{Related Work}
Deep Q-Networks (DQNs) have undergone extensive examination and utilization across diverse domains. The initial endeavour to comprehensively comprehend the DQN algorithm from algorithmic and statistical vantage points was undertaken by Fan et al. \cite{fan2020theoretical}. They rigorously established the rates of convergence, both algorithmically and statistically, pertaining to the action-value functions of the iterative policy sequence derived through the application of the DQN methodology. There have been several extensions to the original DQN algorithm. A previous study by Xiong et al. proposed a parametrized deep Q-network (P-DQN) for addressing hybrid action spaces that encompass both discrete and continuous dimensions \cite{xiong2018parametrized}. Notably, this framework refrains from employing any approximation or relaxation techniques. Another study by Van Hasselt et al. is the Double Deep Q-Network (Double DQN) \cite{van2016deep}. The proposed approach employs Double Q-learning to mitigate the issue of overestimation by decomposing the max operation within the target function. Specifically, this decomposition involves separating the action selection and action evaluation processes.

The study conducted by Ronecker et al. presents an approach that integrates deep Q-networks and control theory to improve decision-making in autonomous driving \cite{ronecker2019deep}. The proposed method aims to ensure the safe navigation of autonomous vehicles in highway scenarios. The researchers conducted an experiment using a Deep Q-Network in a simulated environment to function as a central decision-making entity. They propose targets for a trajectory planner and utilize the resulting trajectories, along with a controller for longitudinal movement, to execute lane change manoeuvres. The authors assess the effectiveness of their methodology in two distinct highway traffic scenarios and analyze the influence of various state representations on performance and training. The findings indicate that the proposed system is capable of generating driving behaviour that is both efficient and safe.

Another research article introduces a decision-making framework for virtual agents in real-time, which is based on the Deep Q-Network (DQN) approach. The framework is designed to automatically select goals for virtual agents in virtual simulation environments, taking into account their motivations, and subsequently plan appropriate behaviours to achieve those goals. The study categorizes all motivations according to Maslow's hierarchy of needs, which consists of five levels. The virtual agents are trained using a double DQN algorithm, utilizing a large dataset of social data. They then select optimal goals based on their motivations and execute behaviours using predefined hierarchical task networks (HTNs). The authors compare their proposed framework with the current state-of-the-art method and observe that it is highly efficient, resulting in a significant reduction in average loss from 0.1239 to 0.0491 and an increase in accuracy from 63.24\% to 80.15\% \cite{jang2021deep}.

A study by Perera et al. discusses the potential of reinforcement learning to optimize energy systems and promote the adoption of renewable energy; the authors highlight the challenges posed by the increasing complexity of energy systems, including the integration of renewable energy technologies, energy storage, and dispatchable energy sources, they discuss how reinforcement learning can be used to address these challenges by providing intelligent and flexible control and optimization of energy systems \cite{perera2021applications}.

As per our knowledge, no DQN applications for optimizing PV installations in dairy farming exist to date. Through the utilization of DQNs, agricultural investors can make well-optimized choices that lead to heightened energy efficiency, minimized environmental impact, and increased profitability. 
\section{Background of RL model} 
\par
\begin{figure}[h]
  \centering
  \includegraphics[width=0.75\textwidth]{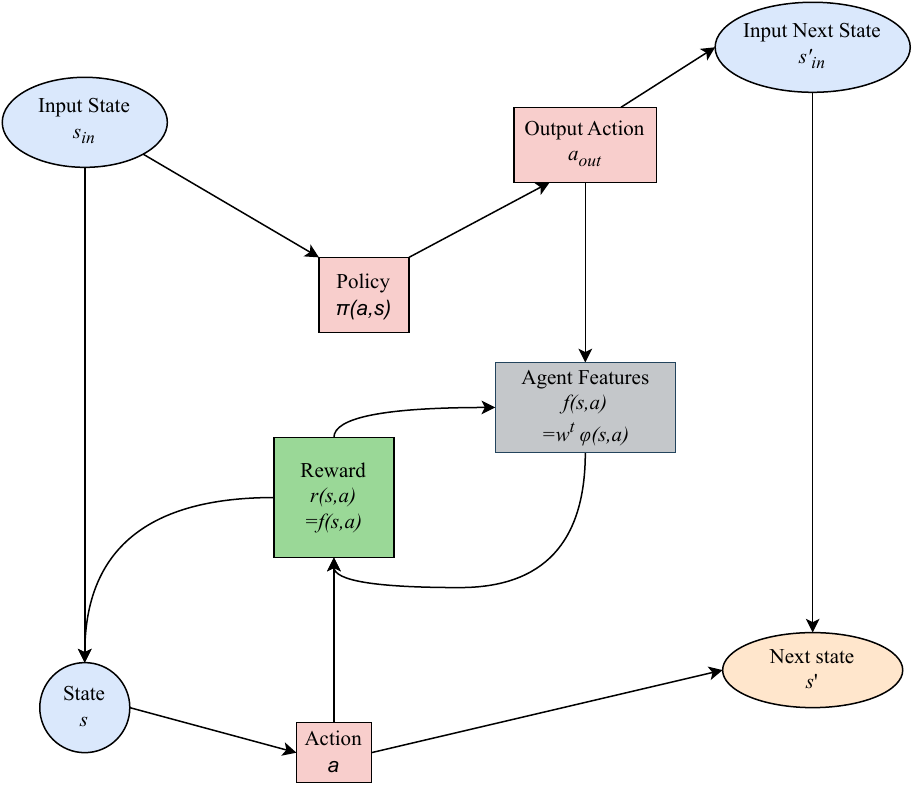}
  \caption{Schematic representation of RL framework.}
  \label{fig:image}
\end{figure}

 Reinforcement Learning (RL) is a specific domain within the field of machine learning that focuses on the acquisition of optimal decision-making abilities by an agent through iterative interactions with an environment, wherein the agent receives feedback in the form of rewards \cite{kaelbling1996reinforcement}. Reinforcement learning (RL) is particularly advantageous when the agent is required to acquire knowledge through experience and sequentially make decisions. The schematic diagram of an RL framework is depicted in Figure \ref{fig:image}.
\par The mathematical model of Reinforcement Learning (RL) is constructed based on the theoretical framework of Markov Decision Processes (MDPs). MDPs offer a formal mathematical representation for modelling problems involving sequential decision-making \cite{van2012reinforcement,li2022deep}. An MDP is formally represented by a tuple ($S, A, P, R, \gamma$), where: 
\begin{itemize}
    \item The symbol $S$ denotes the set of states.
    \item The symbol $A$ denotes the set of actions available to the RL agent.
    \item $P(s_t, a_t, s_{t+1})$ represents the state-transition probability function. 
    \item The reward function $R(s_t, a_t, s_{t+1})$ assigns an immediate reward $r_t$ to the agent when it transitions from state $s_t$ to $s_{t+1}$ by taking action $a \in A$. 
    \item The discount factor $\gamma$ (where $0 \leq \gamma \leq 1$) determines the relative significance of future rewards compared to immediate rewards.
\end{itemize}
The primary objective of reinforcement learning (RL) is to identify the optimal policy $\pi^*$ that maximizes the expected cumulative reward over a given time horizon. The cumulative reward, denoted as $R_{cum}$, is computed by summing the discounted rewards, which are defined as:
\begin{equation}
    R_{cum} = \sum_{t=0}^{\infty} \gamma^t R(s_t, a_t, s_{t+1})
    \label{Eqn1}
\end{equation}
Here, the variable $s_t$ is used to represent the current state, $a_t$ is employed to denote the action taken in state $s_t$, and $s_{t+1}$ is utilized to signify the resulting state that takes place after executing action $a_t$. The inclusion of the discount factor $\gamma$ serves to allocate relatively lowered importance to future rewards in comparison to immediate rewards.

\par The determination of the optimal policy $\pi^*$ is achieved by resolving the reinforcement learning problem through iterative interactions between the reinforcement learning agent and the environment. The process entails iteratively estimating the values or Q-values of state-action pairs, subsequently updating the policy based on these estimates, and further refining the estimates through a combination of exploration and exploitation. Nevertheless, the agent frequently lacks knowledge about the dynamics of the environment, encompassing the state-transition probabilities denoted as $P(s_t, s_{t+1}, a_t)$, as well as the reward function represented by $R(s_t, s_{t+1}, a_t)$. Consequently, the agent acquires the most favourable course of action exclusively by engaging with the environment and employing designated learning algorithms.

\par The policy $\pi^*$ that maximizes the expected cumulative reward corresponds to the state-value function $V_{\pi^*}(s)$, which is greater than or equal to the value functions associated with all alternative policies for all states. However, the direct acquisition of the optimal policy $\pi^*$ from the value function $V_{s}$ poses a challenge due to the absence of explicit action information. To tackle this issue, the action-value function $q_{\pi}(s_t, a_t)$, which is commonly referred to as the state-action value function, is introduced within the field of reinforcement learning (RL) \cite{sutton2018reinforcement}. The action-value function is a mathematical representation of the anticipated total future reward that can be obtained by starting from a given state $s$, selecting a particular action $a$, and subsequently adhering to a specific policy $\pi$. Through the utilization of the optimal action-value function $q_{\pi}(s_t, a_t)$, the determination of the optimal policy $\pi^*$ can be achieved by selecting the action that maximizes the value, which is defined as:
\begin{equation}
    a_t= argmax_{a \in A} (q_{\pi*}(s_t, a_t))
    \label{Eqn2}
\end{equation}
\par By employing iterative updates that depend on the Bellman equation, value-based methods such as Q-learning can acquire the optimal action-value function denoted as $q^*$. The Bellman equation, as described by Thomas \cite{thomas2007markov}, establishes a mathematical relationship between the action value of the current state and the sum of the immediate reward and the discounted action value of the subsequent state. Through a process of iterative estimation and updating of action values, value-based techniques can converge towards the optimal action-value function, denoted as $q^*$, and subsequently, towards the optimal policy. The process can be succinctly summarized in this way:
\begin{equation}
    Q_{t+1}(s_t, a_t)= Q_t(s_t, a_t)+\eta_Q[r_t+\gamma \cdot max_a Q_t(s_{t+1}, a)-Q_t(s_t, a_t)]
    \label{Eqn3}
\end{equation}
The expression $Q_{t+1}(s_t, a_t)$ denotes the updated estimation of the action value associated with state $s$ and action $a$ during iteration $t$. The function $Q_t(s_t, a_t)$ represents the previous estimate, while $r_t$ denotes the immediate reward acquired following the execution of action $a$ in state $s$. The symbol $s_{t+1}$ represents the subsequent state in a sequence. The discount factor $\gamma$ is a parameter that determines the relative importance of future rewards. The expression $max_a Q_t(s_{t+1}, a)$ denotes the maximum action value for the next state $s_{t+1}$, considering all possible actions $a$ at the current iteration $t$.
\section{DQN for PV Integration into Agriculture}
\begin{figure}[h]
  \centering
  \includegraphics[width=0.75\textwidth]{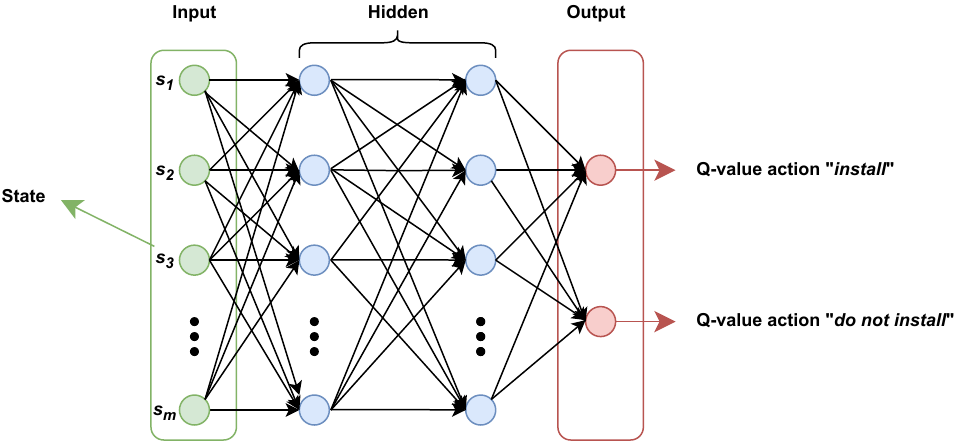}
  \caption{DQN structure.}
  \label{fig:image2}
\end{figure}

Deep Q-network (DQN) is an advanced reinforcement learning algorithm that leverages the power of Deep Neural Networks (DNNs) to establish connections between states and actions, similar to the Q-Table used in Q-Learning \cite{mnih2015human}. DNNs are capable of learning abstract representations directly from raw sensor data. Similar to a Q-Learning agent, a DQN agent interacts with the environment through a sequence of observations, actions, and rewards. The general structure of a DQN is depicted in Figure \ref{fig:image2}, where the network takes the state as input and produces Q-values for each action in the action space. The neural network's primary objective is to train and learn its parameters. During the prediction phase, the trained network determines the optimal action based on the subsequent environmental state.
In the proposed approach, experience replay is used to select experiences from memory while improving the DQN agent randomly. The DQN employs a Deep Neural Network (DNN) as a function approximator with variable weights. The Q-Network is trained by iteratively updating these parameters in each iteration, aiming to minimize the mean squared error in the Bellman equation. The loss function, defined by Equation \ref{Eqn4}, measures the squared difference between the target Q and the predicted Q as follows:
\begin{equation}
    Loss = (r_t + \gamma ^{max}_{a_{t+1}} Q(s_{t+1},a_{t+1};\theta_{t+1})- Q(s_t,a_t;\theta_t))^2
    \label{Eqn4}
\end{equation}
\par The proposed method involves defining the state space, action space, and reward function and training an agent using DQN. The state space includes all the relevant variables that could influence the decision to install solar PV systems, such as electricity usage, the cost of electricity, the amount of sunlight the farm receives, the cost of PV installation, etc. The action space includes all the possible actions that the agent can take in response to a given state, such as installing PV systems or do not install. The reward function outlines the goal that the agent is attempting to achieve, including the price of the solar panels themselves, the cost of the installation, and any other pertinent details.

In the PV context, the deep RL-based decision support system is trained with a specific DNN input representation for the RL state, designated as $s_t$. The state $s_t$ in this particular case study encompasses three variables: the farm load, government incentives, and installation budget. It can be represented as follows: $s = [farm_{load}, Incentives, Budget]$. 

This study emphasizes its current state and action rather than explicitly incorporating future prognostications, as the future is susceptible to significant influence from current events and undertaken actions. To streamline the complexities of the decision-making context, it is postulated that the action $a_t$ exerts an instantaneous influence on the subsequent information available.
The reward function $R_t$ is defined as: 
\begin{equation}
    R_t = - E_p \times (load - PV power)
\end{equation}
Here, the term "$E_p$" represents farm electricity prices, the second term reflects the total farm load, and the last term is the power generated by the PV system.
\par During the training phase, the agent actively engages with the environment, executing actions and observing the subsequent state and associated reward. Over time, the agent learns to improve decision-making by leveraging a neural network to approximate the optimal action-value function. This function estimates the expected reward for taking a specific action in a particular state. By acquiring this knowledge, the agent becomes equipped to provide personalized recommendations to homeowners regarding the installation of solar panels, considering their unique circumstances and requirements. Algorithm 1 describes the training procedure of the DQN agent. 

\par In each learning iteration, DQN utilizes an $\epsilon$-greedy strategy \cite{mnih2015human}. This strategy encourages the agent to randomly choose actions with a decreasing probability $\epsilon$, enabling the exploration of different possibilities during the early stages of learning. This approach facilitates the agent's exploration and learning from diverse actions and their outcomes.

\begin{algorithm}[H]
    \SetAlgoLined
    \SetKwInOut{Input}{Input}
    \SetKwInOut{Output}{Output}
    Initialize an MLP network with random weights $\theta$ \\
    Create an empty replay memory buffer $D$ with a capacity of $N$ \\
    Set current time step $t = 0$ \\
    \For{each episode $e$ from $1$ to $E$}{
        Reset the environment and obtain the initial state $s$ \\
        Set total episode reward $R = 0$ \\
        \While{episode is not terminated}{
            \If{with probability $\epsilon$}{
                Select a random action $a$ from the action space $A$ \\
            }
            \Else{
                Select $a_t = \text{argmax}_a \, Q(s_t, a; \theta_t)$ where $\theta_t$ represents the parameters of the MLP network \\
            }
            Execute action $a_t$ in the environment and observe reward $r_t$ and next state $s_{t+1}$ \\
            Store transition $(s_t, a_t, r_t, s_{t+1})$ in replay memory buffer $D$ \\
            Sample a mini-batch of $B$ transitions $(s_i, a_i, r_i, s_{i+1})$ from $D$ \\
            Compute target Q-values for each transition: \\
            $Q_{\text{target}} = r_i + \gamma \cdot \max_a Q(s_{i+1}, a; \theta_{i+1})$ \\
            Update the weights of the MLP network using gradient descent: \\
            $\theta = \theta - \alpha \cdot \nabla_{\theta} Q(s_i, a_i; \theta) \cdot (Q(s_i, a_i; \theta) - Q_{\text{target}})$ \\
            Update the current state $s = s'$ \\
            Update the total episode reward $R = R + r_t$ \\
            Increment the current time step $t = t + 1$ \\
        }
    }
    \caption{Agent training process of Deep Q Learning.}
\end{algorithm}

\section{Experiment}
This section provides an overview of the experimental analysis and findings from our study. The deep reinforcement learning (RL) decision support system was trained using the hyperparameters outlined in Table \ref{tab:parameter}. The exploration probability, denoted as $\epsilon_t$, was crucial in determining the learning performance. It was configured to decrease exponentially as the iteration number $n_i$ increased, following the equation $\epsilon_t = 0.99999^{n_i}$. However, a minimum value of 0.01 was set as the lower bound for $\epsilon_t$. The training procedure was executed on a Google Colab platform utilizing a GPU (NVIDIA Tesla T4). The process of training, consisting of 500 episodes, required approximately 137 minutes to reach convergence.
\subsection{Hyperparameters}
Table \ref{tab:parameter} presents the hyperparameters of the DQN-based decision-making system used in this study.
\begin{table}[h]
\centering
\caption{Hyperparameters of DQN-based decision-making system.}
\begin{tabular}{|l|l|}
\hline
\textbf{Hyperparameters} & \textbf{Values} \\
\hline
Batch size & 128 \\
Exploration probability ($\epsilon_t$) & $0.99999^{n_i}$ \\
Number of layers & 2\\
Number of neurons of each layer & 32\\
Learning rate & 0.1\\
Activation functions in hidden layers & ReLU\\
\hline
\end{tabular}
    \label{tab:parameter}
\end{table}
\subsection{Results and Analysis}
\begin{figure}[h]
\centering
\includegraphics[width=0.74\textwidth]{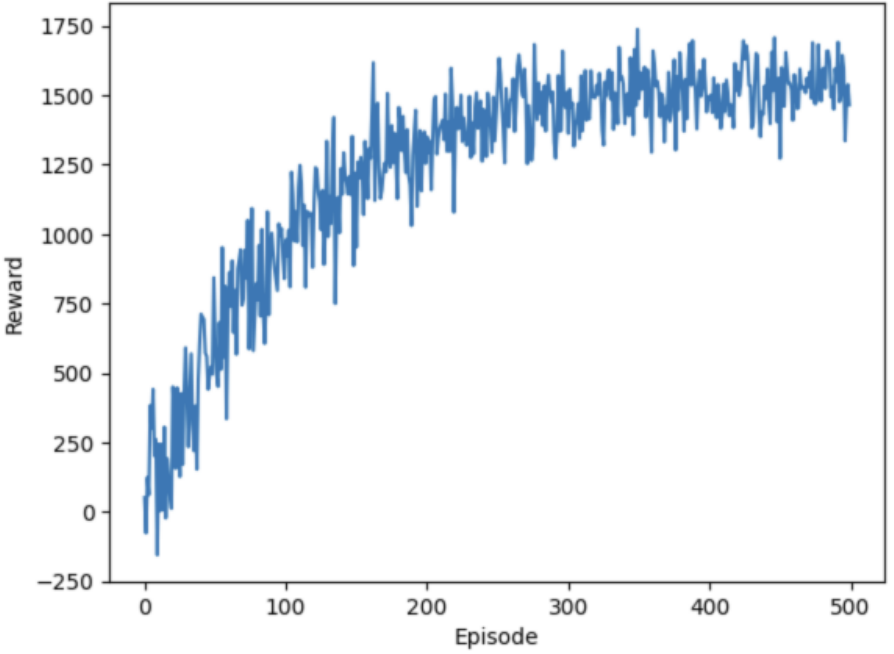}
\caption{Reward per episode.}
\label{fig:dqn}
\end{figure}

The performance of deep Q learning is depicted in Figure \ref{fig:dqn}, which shows the instantaneous reward achieved per episode. The learning curve demonstrates significant variation even when the exploration probability is set to its minimum value of 0.01. The observed variation can be linked to the stochastic nature of the actions taken within the learning environment during each episode rather than indicating any inherent instability in the learning process. This model performed well in determining optimal actions for PV system investment decisions, including installation and don't install scenarios. Throughout the training process, the model consistently improved its decision-making capabilities, as evident from the learning curve in Figure \ref{fig:dqn}. Converging at 500 epochs, the model reached a stable state, ensuring consistent and accurate predictions. This level of accuracy is an indicator of the model's ability to make appropriate decisions based on budget constraints and farm electricity usage data.

\par
\begin{figure}[h]
\centering
\includegraphics[width=0.74\textwidth]{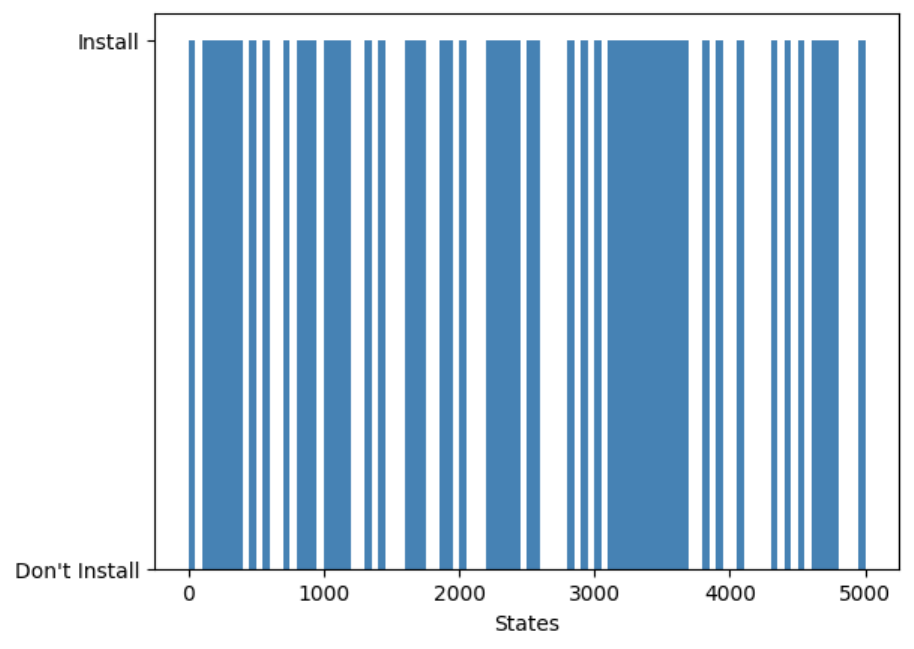}
\caption{DQN decision over states.}
\label{fig:dqn_decision}
\end{figure}

 We have conducted a performance analysis of our DQN model's performance and outlined its decision-making process for installing solar photovoltaic (PV) systems in various states. In Figure \ref{fig:dqn_decision}, we can observe the outcomes of the DQN model's decisions, specifically whether to recommend the installation or not of PV systems. Each vertical bar within the graph corresponds to a particular state, representing the model's recommendation. By analyzing the distribution of bars, we can identify patterns in the model's recommendations. Clusters of blue bars indicate regions where the model consistently suggests installation, while other groupings indicate regions where it consistently advises against installing PV systems.\\
For the "Don't Install" Decision:
\begin{itemize}
    \item \textbf{Farm Load}: Instances with comparatively lower farm loads exhibit a propensity for the "Don't Install" recommendation. These situations likely indicate scenarios where the energy demand is modest, potentially rendering the implementation of a solar PV system less advantageous.
    \item \textbf{Budget and System Cost}: The model recommends against installation when the budget falls notably short of covering the system cost. This fiscally conservative stance aligns with instances where financial considerations undermine the viability of solar PV deployment.
\end{itemize}
For the "Install" Decision:
\begin{itemize}
    \item \textbf{Farm Load}: Enhanced farm loads are crucial in promoting the "Install" recommendation. Elevated energy demands, indicative of significant load requirements, tend to elicit a favourable inclination towards system installation.
    \item \textbf{Budget and System Cost}: The availability of a sufficiently robust budget to meet or exceed the system cost is pivotal in steering the model towards the "Install" recommendation. In cases where the budget accommodates the financial implications, the model is more likely to endorse solar PV deployment.
\end{itemize}
\par
\begin{figure}[h]
     \centering
     \begin{subfigure}[b]{0.49\textwidth}
         \centering
         \includegraphics[width=\textwidth]{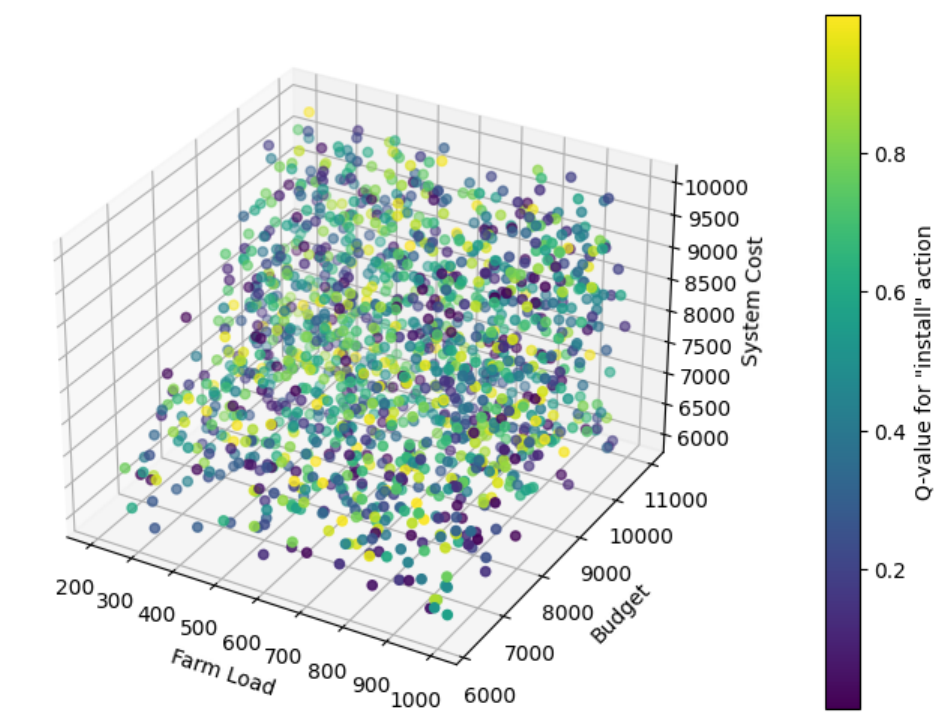}
         \label{fig:install}
     \end{subfigure}
     \hfill
     \begin{subfigure}[b]{0.49\textwidth}
         \centering
         \includegraphics[width=\textwidth]{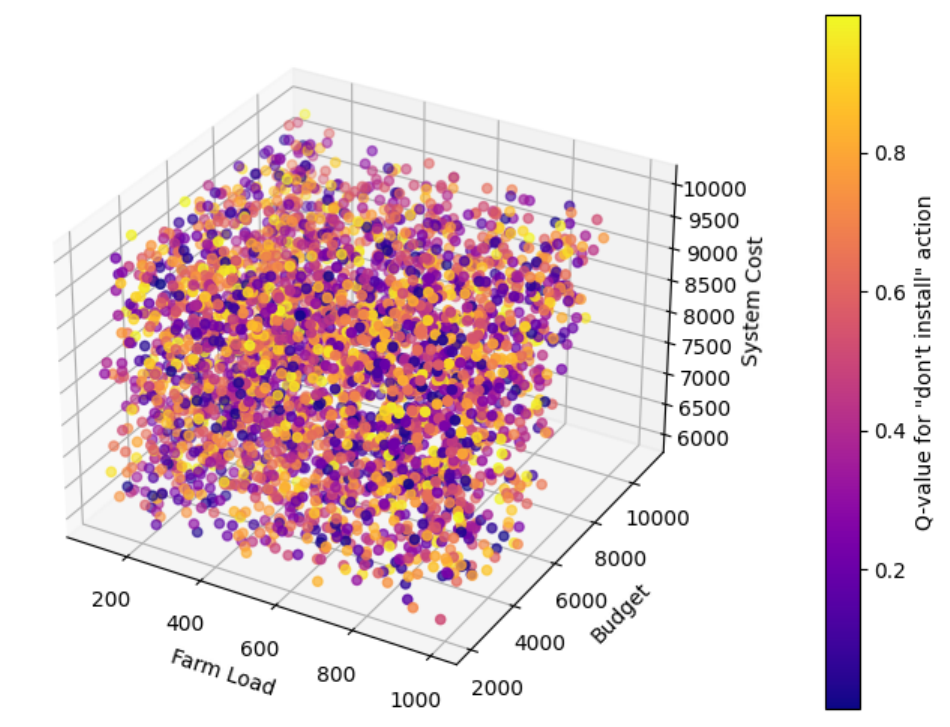}
         \label{fig:donot_install}
     \end{subfigure}
        \caption{Q-values attributed to the  (a) "Install" and (b) "Don't Install" action.}
        \label{fig:decision}
\end{figure}

 Further, we have analyzed how these state variables are mapped to Q-values, representing the anticipated cumulative rewards associated with each action. The analysis is presented through two distinct 3D plots. In Figure \ref{fig:decision}(a), the Q-values corresponding to the "Install" action are illustrated, employing a spectrum of colours ranging from cooler to warmer tones. Here, warmer colours indicate a higher propensity for recommending system installation, while cooler colours suggest a less favourable stance.  Figure \ref{fig:decision}(b) mirrors this approach but focuses on the Q-values for the "Don't Install" action. We can observe that in regions where installing solar PV is economically and practically viable (for example, higher budget, lower system cost, higher farm load), the Q-values for the "Install" action is high. Similarly, in regions where installation is not advantageous, the Q-values for the "Don't Install" action are high. These plots allow us to visually identify where the DQN model suggests installing or not installing the solar PV system based on the state variables. By analysing these plots, decision-makers can readily discern regions where solar PV installation is advantageous. The model's efficacy hinges upon identifiable patterns within Q-values, thereby underscoring the DQN's capacity to encapsulate the intricate interplay of factors that shape decisions about deploying solar PV systems.
\section{Conclusion and Future Work}

The research findings demonstrate the effectiveness of integrating PV systems in the farm. The experimental results reveal significant reductions in energy costs and reliance on the conventional power grid through the integration of PV in farm operations. The main findings of this research are: 
\begin{enumerate}
    \item It was demonstrated that by integrating PV systems, farms decrease their reliance on the utility grid, reducing energy purchases.
    \item The integration of PV systems enabled farms to lower their energy expenses, leading to long-term cost savings. This positively impacts the overall financial performance of agricultural operations.
    \item This research contributed to the understanding of the economic implications associated with the adoption of renewable energy solutions in the farming sector.
\end{enumerate}
Future work in this area could explore using more complex state and action spaces to capture additional factors influencing solar panel installation decisions. Additionally, alternative reinforcement learning algorithms could be investigated to enhance the recommendation capabilities further. 
\section*{Acknowledgements} This publication has emanated from research conducted with the financial support of Science Foundation Ireland under Grant number [21/FFP-A/9040].

\bibliography{template}
\end{document}